\documentclass{article}

\usepackage{PRIMEarxiv}

\usepackage[utf8]{inputenc} 
\usepackage[T1]{fontenc}    
\usepackage{hyperref}       
\usepackage{url}            
\usepackage{booktabs}       
\usepackage{amsfonts}       
\usepackage{nicefrac}       
\usepackage{microtype}      
\usepackage{lipsum}
\usepackage{fancyhdr}       
\usepackage{graphicx}       
\graphicspath{{media/}}     
\usepackage{float}
\usepackage{subfig}
\title{Generative Adversarial Networks for anonymous 
Acneic face dataset generation
\thanks{\textit{\underline{Citation}}: 
\textbf{Authors. Title. Pages.... DOI:000000/11111.}} 
}

\author{
  Hazem Zein \\
  LISSI Laboratory \\
  Université Paris-Est Créteil \\
  F-94400 Vitry-sur-Seine, France\\
  \texttt{hazem.zein@u-pec.fr} \\
   \And
  Samer Chantaf \\
  Faculty of Technology \\
  Lebanese University \\
  Saida, Lebanon\\
  \texttt{samerchantaf@gmail.com} \\
   \And
  Régis Fournier \\
  LISSI Laboratory\\
  Université Paris-Est Créteil \\
  F-94400 Vitry-sur-Seine, France\\
  \texttt{rfournier@u-pec.fr} \\
     \And
  Amine Nait-Ali \\
  LISSI Laboratory \\
  Université Paris-Est Créteil \\
  F-94400 Vitry-sur-Seine, France\\
  \texttt{naitali@u-pec.fr} \\
}

\begin{document}
\maketitle

\begin{abstract}
It is well known that the performance of any classification model is effective if the dataset used for the training process and the test process satisfy some specific requirements. In other words, the more the dataset size is large, balanced, and representative, the more one can trust the proposed model's effectiveness and, consequently, the obtained results. Unfortunately, large-size anonymous datasets are generally not publicly available in biomedical applications, especially those dealing with pathological human face images. This concern makes using deep-learning-based approaches challenging to deploy and difficult to reproduce or verify some published results. In this paper, we suggest an efficient method to generate a realistic anonymous synthetic dataset of human faces with the attributes of acne disorders corresponding to three levels of severity (i.e. Mild, Moderate and Severe). Therefore, a specific hierarchy StyleGAN-based algorithm trained at distinct levels is considered. To evaluate the performance of the proposed scheme, we consider a CNN-based classification system, trained using the generated synthetic acneic face images and tested using authentic face images. Consequently, we show that an accuracy of 97,6\% is achieved using InceptionResNetv2.
As a result, this work allows the scientific community to employ the generated synthetic dataset for any data processing application without restrictions on legal or ethical concerns. Moreover, this approach can also be extended to other applications requiring the generation of synthetic medical images. We can make the code and the generated dataset accessible for the scientific community.
\end{abstract}

\keywords{Convolutional Neural Networks \and Deep learning \and Facial acne disease \and Generative Adversarial Networks}

\section{Introduction}
\label{sec:introduction}
Skin diseases occur in people of different ages and cultures and they are more common than other diseases worldwide \cite{b0} and can be caused by many factors such as diet, hormones, virus, and bacteria. There are various types of skin diseases like acne, eczema, rosacea, psoriasis, vitiligo... A study conducted in 2017 showed that dermatologists density in the United States is 3.4 per 100000 persons which is still lower than the needed density \cite{b00}. Deep Learning provides the ability to improve the work of dermatologist with an AI assisted diagnosis system which lowers the  burden on dermatologists. Deep Learning has been proven to yield great results on many image classification tasks including dermatology applications, and it can help classify skin diseases that are highly similar to each other, but in the dermatology field we face multiple obstacles before collecting data needed to feed deep learning models such as unavailability of public datasets, privacy and legal issues, low quality datasets: datasets focused only on some types of skin diseases. Generative adversarial networks are the solution to skip all these obstacles and present a way to augment our data with artificial images. GANs belong to the generative modeling type. GANs learn to generate synthetic samples indistinguishable from real ones after being trained on real samples. Many published research papers have used GANs in the generation of skin disease images and CNNs to classify skin disease types. 
In 2019 Baur et. Al. used DCGAN and LAPGAN on ISIC2018 dataset to generate synthetic images of skin lesions \cite{b1}.
In 2019 Ghorbani et Al. proposed a model based on Pix2Pix architecture to generate images of skin lesions with a resolution of 256x256. Their dataset consisted of 49920 high resolution images \cite{b2}. Xiang et. Al. in 2020 trained AC-GAN using HAM10000 containing skin lesions images with a resolution of 64x64 upscaled to 256x256  to train multiple classification models for result comparison \cite{b5}. Junayed et. Al. proposed a CNN architecture “AcneNet” which was trained on a dataset of 5 classes of acne containing 360 images per class with a resolution of 224x224. AcneNet reached an accuracy of 99.44\% in one class and a minimum of 94\% for the rest of the classes\cite{b3}.
Wu et. al. (2019) trained different CNN models on a dataset of 6 classes of skin diseases and a total of 2656 images. The best model achieved 77\% average precision \cite{b4}. In 2021, Srinivasu et. al. used MobileNetV2 and LSTM that were trained on HAM10000 skin lesions dataset and they achieved an accuracy of 85\% \cite{b51} . In this paper we propose using generative adversarial networks to give the ability to the medical community to generate datasets of realistic artificial faces with acne disease and to remove the barriers on access to data by generating highly realistic synthetic images that can allow us to reach same results as when using real images. To prove that the generated images can replace real images in deep learning applications, we trained three convolutional neural networks: InceptionResNetV2, ResNet50V2 and ResNet152V2 only on synthetic images and testing was done on real facial acne images not included in the training process. This paper is structured as follows: In section \ref{materials} we present the methodology including dataset gathering, algorithms and preprocessing steps to reach our objectives. In section \ref{results} we discuss all the results obtained and finally we conclude this paper in Section \ref{conclusion}.

\section{Materials and Methods}

In recent years, deep learning is being implemented more in skin disease applications due to its high accuracy in classifying diseases compared to traditional methods. Subsection \ref{subsection:cnn} and \ref{subsection:gan} give a brief overview on convolutional neural networks and generative adversarial networks and present the different deep learning models used in this paper. In subsection \ref{section:methodology} we discuss the methodology used in this paper.

\label{materials}

\subsection{Convolutional Neural Networks}
\label{subsection:cnn}
Convolutional Neural Networks belong to the class of artificial neural networks and their accuracy has surpassed traditional methods and specifically when the data is images. It is the most popular and mostly used among deep learning algorithms. CNNs are composed of an input layer multiple hidden layers and an output layer and  have multiple types of layers including convolutional layer, pooling layer and fully-connected layer. CNNs are being used in many applications like object recognition, image classification, natural language processing \cite{b61}.
In this Paper we used 3 Pre-Trained CNN Models:
\begin{itemize}
    \item InceptionResNetV2 is a hybrid of Inception networks and Residual Connections without filter concatenation but it comes with a heavier computational cost \cite{b9},
    \item ResNet152V2 and ResNet50V2 are two models with high number of layers as ResNet152V2 has 152 layers and ResNet50V2 has 50 layers. The main difference between ResNet V1 and V2 is that in V2  batch normalization and relu activation are used before each weight layer \cite{b16}.
\end{itemize}

\subsection{Generative Adversarial Networks}
\label{subsection:gan}
\begin{figure}[H]
  
    \center
    \includegraphics[width=\linewidth]{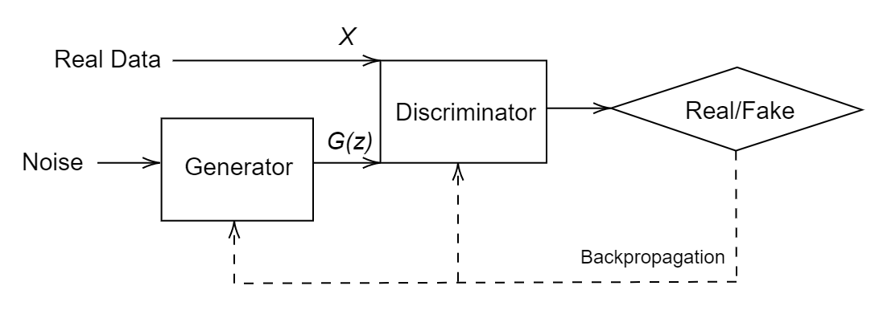}
    \caption{GAN diagram composed of two models: generator and discriminator. Backpropagation is used to minimize the loss of the two models}
    \label{fig:gan-diagram}
  \end{figure}

GANs are composed of two models as shown in figure \ref{fig:gan-diagram}: generator and discriminator. The generator’s role is to generate synthetic samples to fool the discriminator into classifying these samples as real. The discriminator classifies data as either coming from original dataset or from the generator’s output. Both models are competing to improve each other and using backpropagation to reduce each model’s loss by adjusting the weights of each model depending on the impact on the output \cite{b7}.

\subsubsection{StyleGAN2}
StyleGAN developed by NVIDIA in 2018, brings the following improvements to the generator model architecture:
\begin{itemize}
\item Baseline Progressive GAN.
\item Tuning and bilinear up/down sampling
\item Addition of mapping, styles and adaptive instance normalization.
\item Removal of latent vector input to generator.
\item Noise addition at each block of the generator.
\item Addition Mixing regularization.\cite{b13}
\end{itemize}

\begin{figure}[H]
  \includegraphics[width=\linewidth]{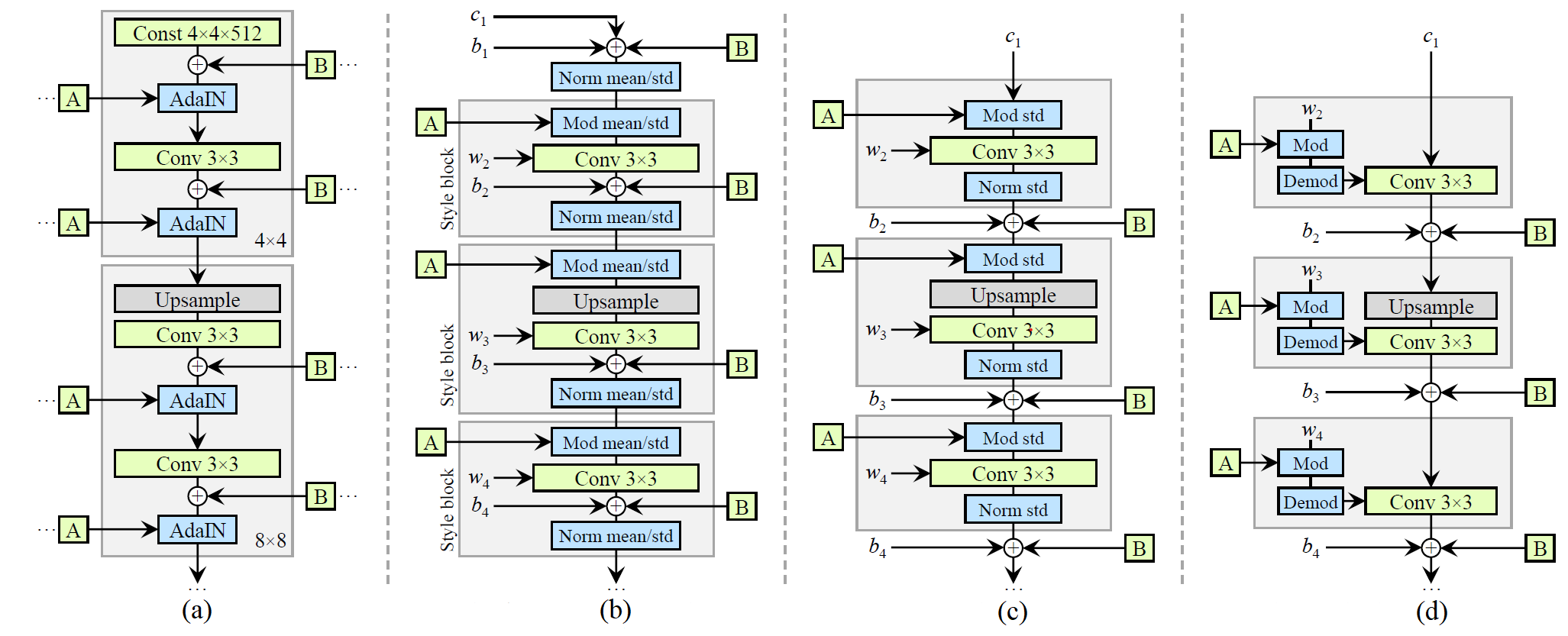}
  \caption{Comparison of StyleGAN and StyleGAN2 architecture, moving from the use of adaptive instance normalization in the StyleGAN to using weight demodulation in StyleGAN2 and shifting the addition of noise to the outside of generator blocks: (a) StyleGAN Architecture. (b) StyleGAN detailed. (c) StyleGAN Revised Architecture. (d) Weight demodulation \cite{b13} \cite{b11}. }
  \label{fig:SA1}
\end{figure}

Figure \ref{fig:SA1} (a) and (b) show the architecture of the first StyleGAN and part (c) is the result of switching to a form that does not use AdaIN (Adaptive Instance Normalization). The same operation as normalization by AdaIN is performed by an operation called Weight demodulation (dividing the weight of the Conv layer by the standard deviation) \cite{b11}.

\subsection{Methodology}
\label{section:methodology}
\begin{figure*}[ht!]
    \centering
    \includegraphics[width=\linewidth]{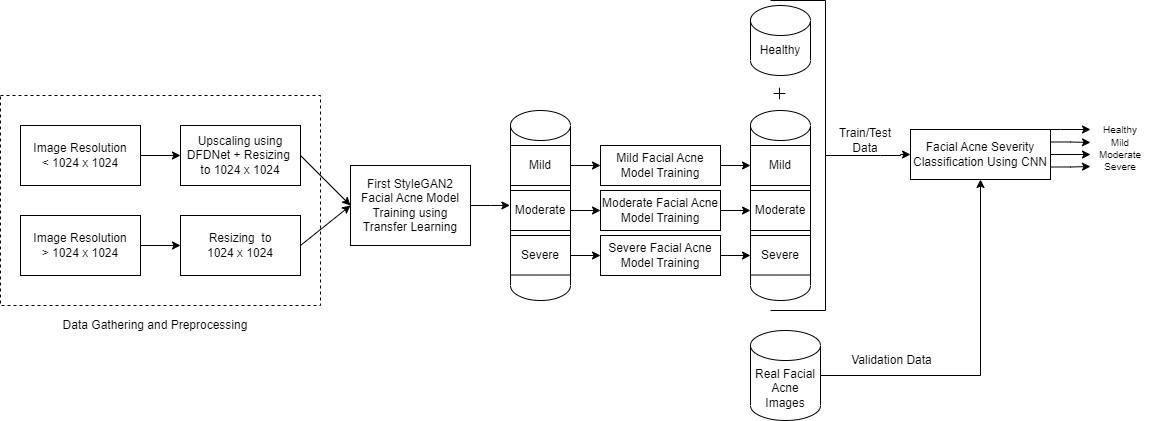}
    \caption{Block Diagram representing the methodology: First step is gathering facial acne images from multiple sources, Second data processing to be compatible with the GAN input layer, then training StyleGAN2 models with different levels of facial acne severity using transfer learning and sorting the output images into 3 classes: mild, moderate and severe and adding a synthetic healthy class. Finally training a CNN model on the previous synthetic dataset to evaluate its performance on real images}
    \label{fig:methodology-diagram}
\end{figure*}

In this paper, we followed the methodology shown in figure \ref{fig:methodology-diagram} to reach our objective. We used style based generative adversarial network 2 (StyleGAN2) to generate our datasets of facial acne disease due to the high resolutions it can reach (1024x1024) and highly realistic results. 
Although we collected 1473 images of different facial acne severities, this dataset was not enough for training 3 GANs to generate each level of severity (Mild, Moderate, Severe).
To overcome this problem, we proceeded with the following steps:
\begin{enumerate}
    \item \textbf{Data Gathering}: First source of images is ACNE04 \cite{b6} dataset. This dataset consists of 4 levels of facial acne severity labeled by dermatologists reaching a total of 1073 images: Low severity(Level 0) with 418 images, Medium severity (level 1) contains 526 images and high severity (Level 2) has 129 images. The resolution of images in the Acne04 is 3112x3456. 400 facial acne images with different levels of severity were gathered from google images with low resolutions. Samples of the 1473 collected images of facial acne are shown in figure \ref{tab:res-images}.
    \item \textbf{Data pre-processing}: because the input images of StyleGAN2 need to be 1024x1024, resizing each input is required. Specifically, if the input is oversized, simple downsampling can be considered. However, if the input image has a  lower resolution, it needs to be upscaled. In this work, we used DFDNet \cite{b8}. From the literature, DFDNet has been successfully employed to unblur and restore human face images. It is known as a deep face dictionary network for face restoration.
    \item \textbf{All Levels of severity acneic face image generation}:
    Using StyleGAN2, this stage aims to generate synthetic acneic face images without distinguishing their severity level. In other words, the output of the StyleGAN2  can be either Mild, Moderate or Severe acneic face images. For this purpose, we consider a model pre-trained on celebrity faces (stylegan2-ffhq-config-f). Afterwards, a transfer learning approach is applied to update the last layer of StyleGAN2, considering the collected images (i.e. 1473 acneic face images) as input data. Transfer learning is widely used in deep learning because it allows training models from small datasets, decreasing the training time and the processing power  \cite{b10}. 
    \item \textbf{Mild/Moderate/Severe acneic face image generation }: the previous StyleGAN2 allows for generating an infinite number of composite synthetic acneic face images. Once sorted into Mild, Moderate and Severe, three distinguished datasets are therefore obtained at this stage. Subsequently, three specific StyleGAN2 models are considered separately. First, the Mild acneic face image dataset is used as an input to a first pre-trained StyleGAN2 (from celebrity faces) to update the network's last layer using transfer learning. This operation is repeated with Mild and Severe acneic face image datasets. Each StyleGAN2 is designed to generate one of the required cases.
    \item \textbf{Hybrid synthetic-authentic classification}: in this stage, the aim is to design a hybrid synthetic-authentic classification system where the training phase is conducted by the synthetic generated images, including healthy face images, whereas the test phase employs authentic acneic face images. In other words, the input dataset merges five sub-datasets. Three of them are acquired from the output of StyleGAN2 models (Mild, Moderate and Severe), and one sub-dataset of health faces is generated with StyleGAN2. Finally, the last sub-dataset (i.e. Test phase purpose) contains only authentic acneic face images.
For comparison and performance evaluation of the classification stage, we considered, in this study, three CNN-based synthetic-authentic classifier systems: InceptionResNetV2, ResNet152V2 and  ResNet50V2.  Four output classes are required: Healthy, Mild, Moderate, Severe and Healthy face images.
\end{enumerate}

\section{Results and Discussion}
\label{results}
We trained all our models on the NVIDIA DGX-1 server in the LISSI laboratory at UPEC. This server contains the following hardware: 
\begin{itemize}
    \item 8x NVIDIA Tesla® V100 16 GB/GPU with 40960 Total NVIDIA CUDA® Cores and 5120 Tensor Cores interconnected with NVIDIA NVLinkTM (1 gpu was used in the training)
    \item 2x 20-Core Intel® Xeon® CPU E5-2698 v4 2.2 GHz 
    \item 512 GB DDR4 LRDIMM Memory
    \item 4X 1.92 TB SSDs RAID 0
\end{itemize}

\subsection{StyleGAN2 Performance Evaluation}
For evaluating our StyleGAN2 models, we used Fréchet Inception Distance as an evaluation metric Fréchet inception distance (FID) was introduced by Heusel et. al. in 2017 \cite{b12}. It calculates the similarity between synthetic and real images at a deep convolutional layer called InceptionV3. It does not compare images at pixel by pixel instead; it calculates the mean and standard deviation and it is defined as:
\begin{equation} 
FID = \left \| \mu_{r} - \mu_{g}  \right \|^{2} +T_{r}(\sum_{r}+\sum_{g} -2(\sum_{r}\sum_{g})^{1/2})
\label{eq1}
\end{equation}
\begin{itemize}
\item $\mu_{r}$ and $\mu_{g}$ real and generated images features' means
\item $\sum_{r}$ and $\sum_{g}$ are the variance matrices.
\item $T_{r}$ is the Trace operator.
\end{itemize}

\begin{figure}[H]
    \centering
    \subfloat{
            \begin{minipage}{0.49\textwidth}
        \centering
        \includegraphics[width=1\linewidth]{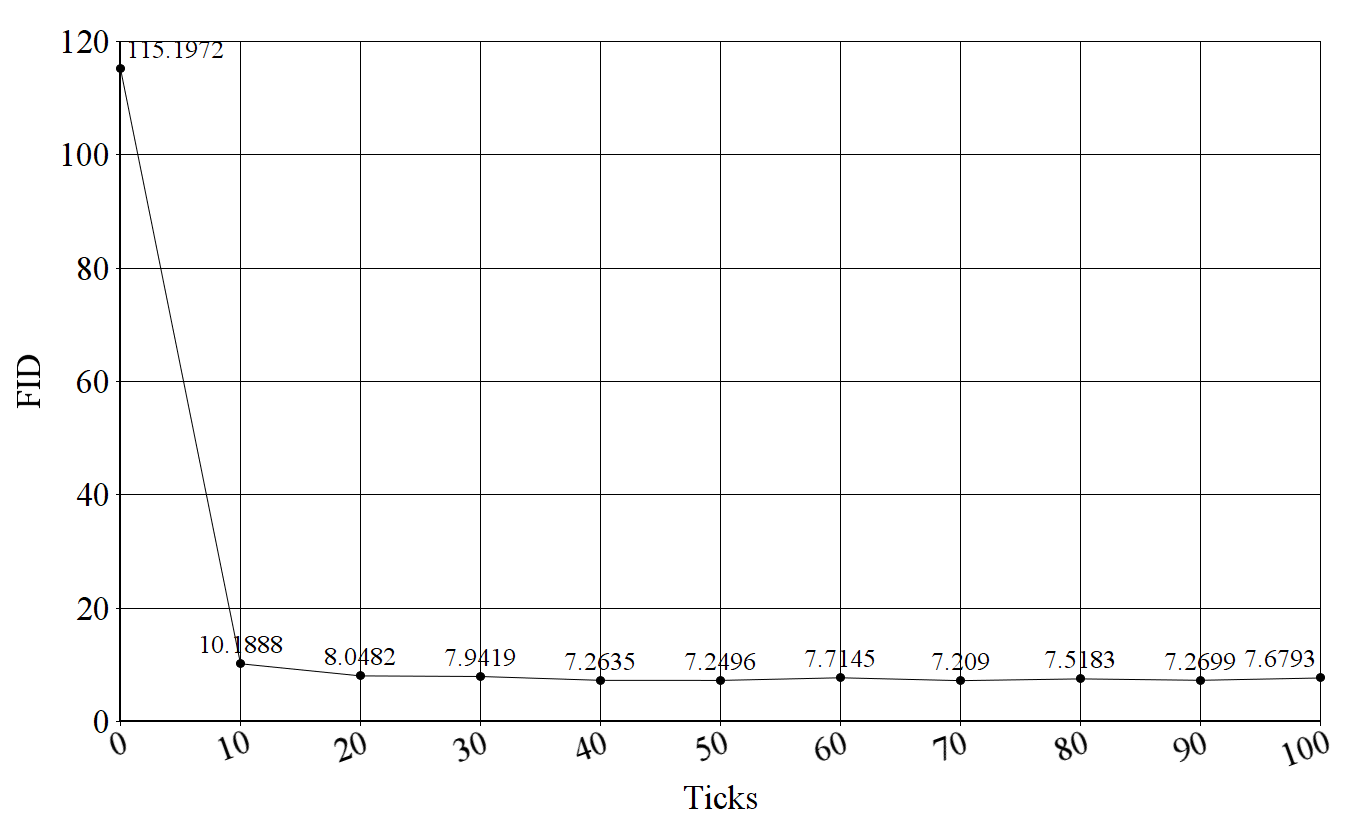}
        (a) \label{fig:first-model}
        \end{minipage}} 
    \subfloat{  \begin{minipage}{0.49\textwidth}
        \centering
        \includegraphics[width=1\linewidth]{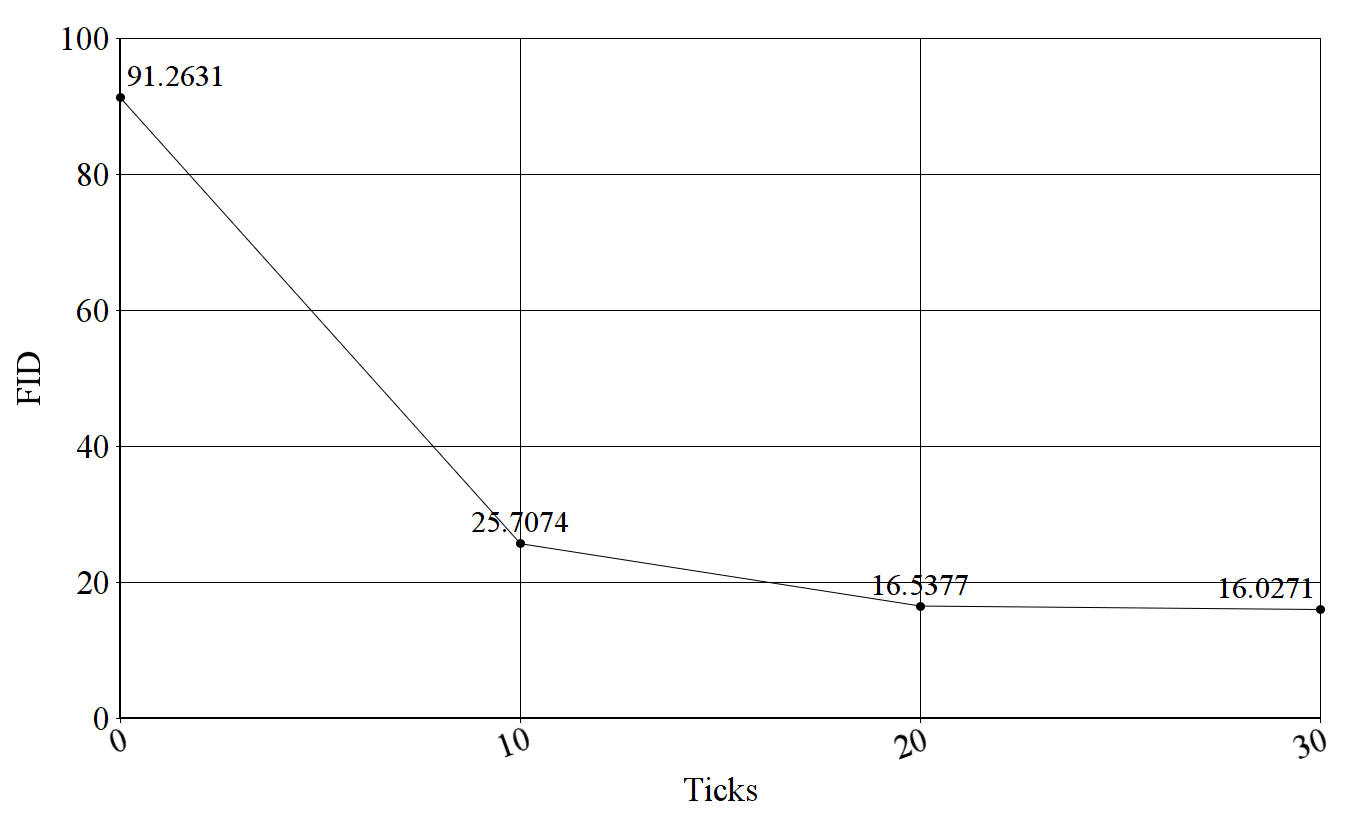}
        (b) \label{fig:first-model}
        \end{minipage}}

    \subfloat{  \begin{minipage}{0.49\textwidth}
        \centering
        \includegraphics[width=1\linewidth]{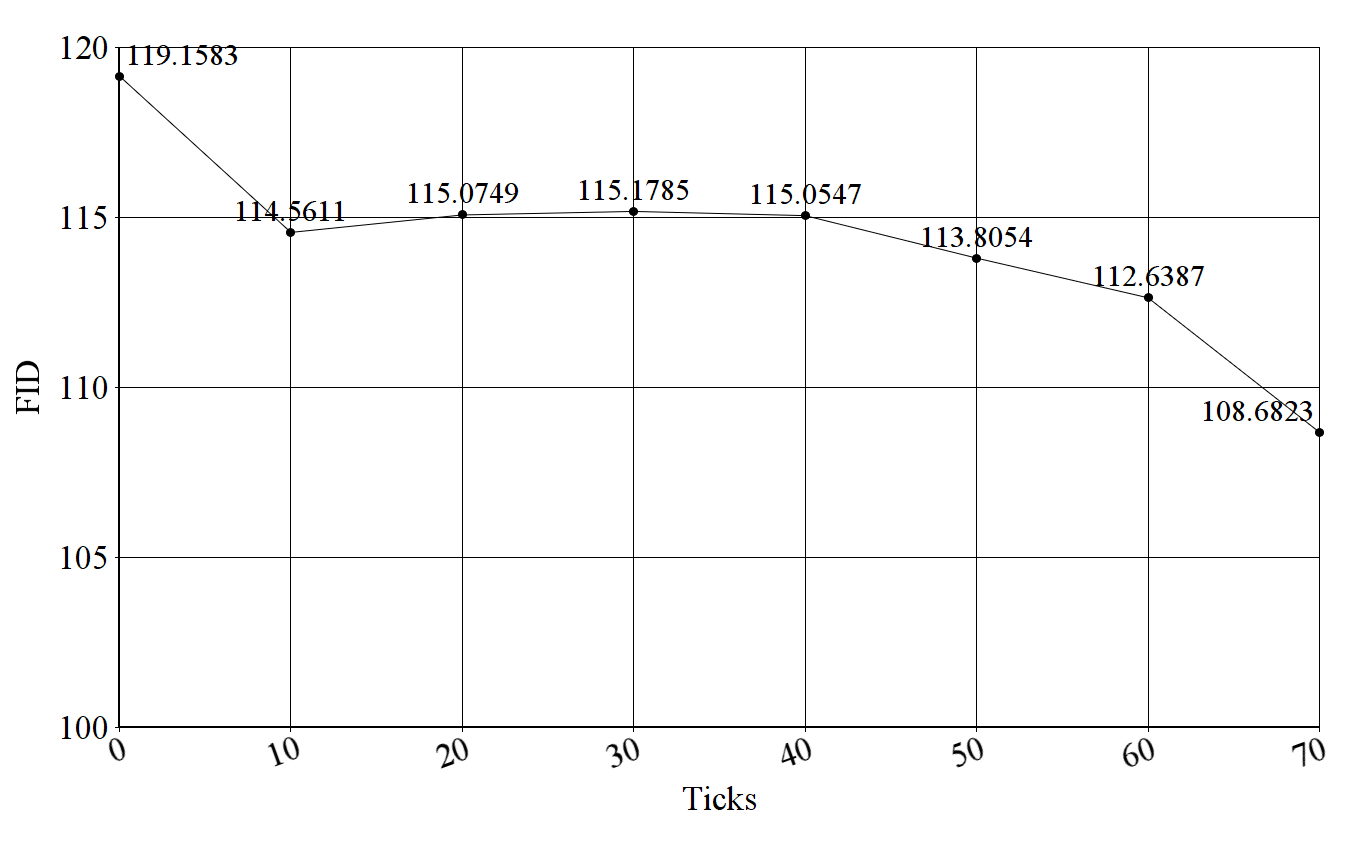}
        (c) \label{fig:first-model}
        \end{minipage}}
    \subfloat{  \begin{minipage}{0.49\textwidth}
        \centering
        \includegraphics[width=1\linewidth]{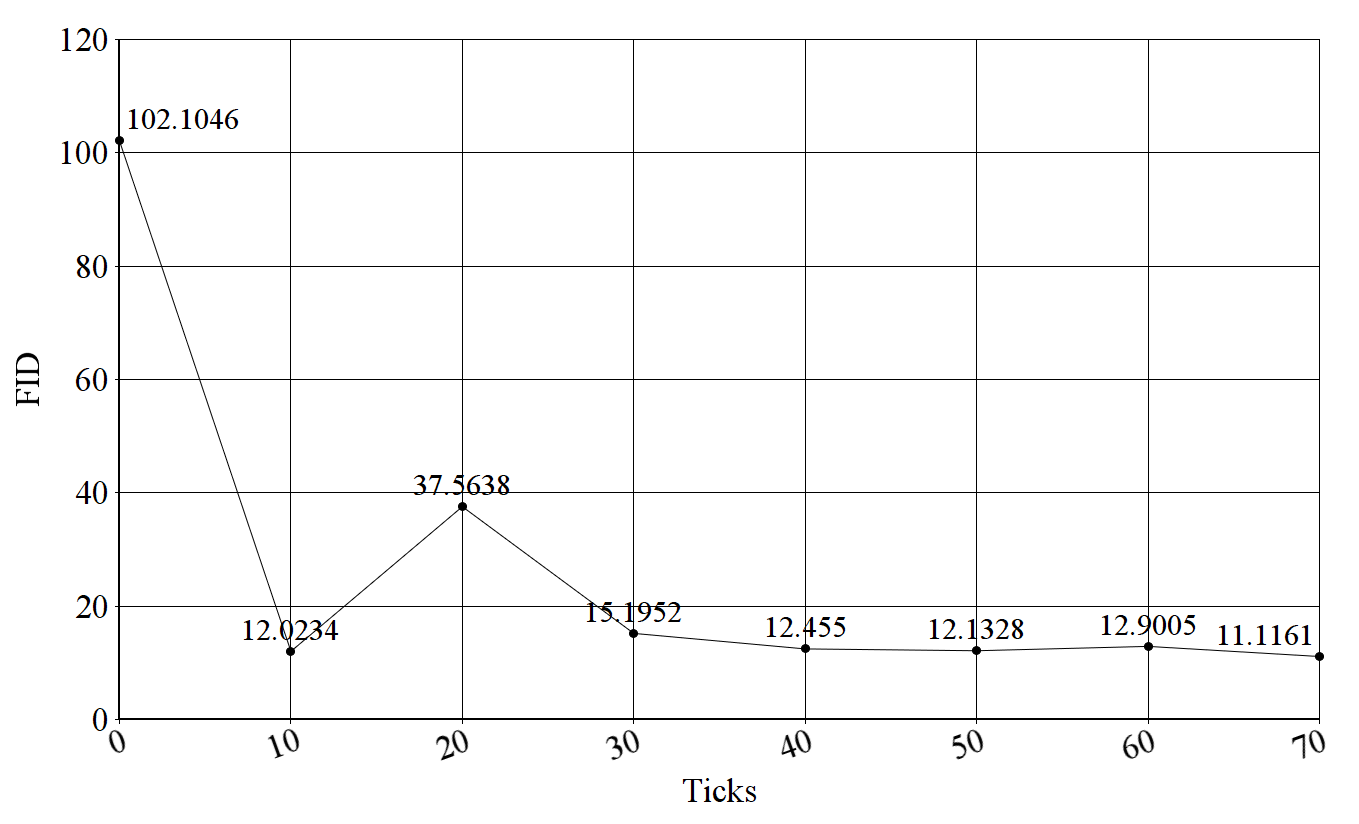}
        (d) \label{fig:first-model}
        \end{minipage}}
\caption{StyleGAN2 models FID graphs: (a) First Facial Acne Model FID Graph, (b) Mild Model FID Graph, (c) Moderate Model FID Graph, (d) Severe Model FID Graph}
    \label{tab:GAN-GRAPHS}
\end{figure}

\begin{figure}[H]
\centering
\resizebox{0.94\textwidth}{!}{%
\begin{tabular}{ccccc}
\small{Real Images} & \small{First Facial Acne Model} & \small{Mild Acne Model} & \small{Moderate Acne Model} & \small{}{Severe Acne Model} \\ 
\includegraphics[height=3cm,width=3cm]{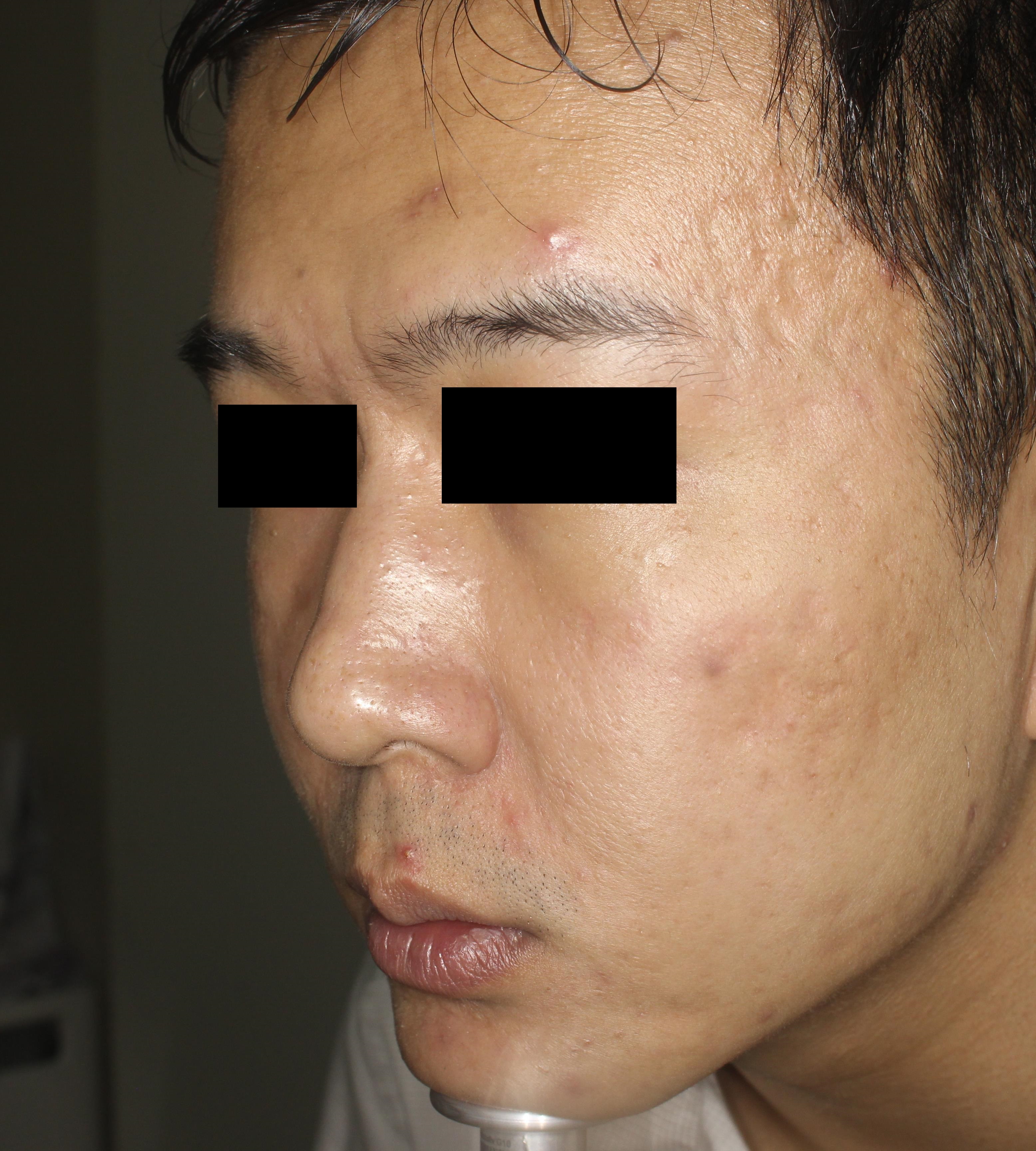}          & 
\includegraphics[height=3cm,width=3cm]{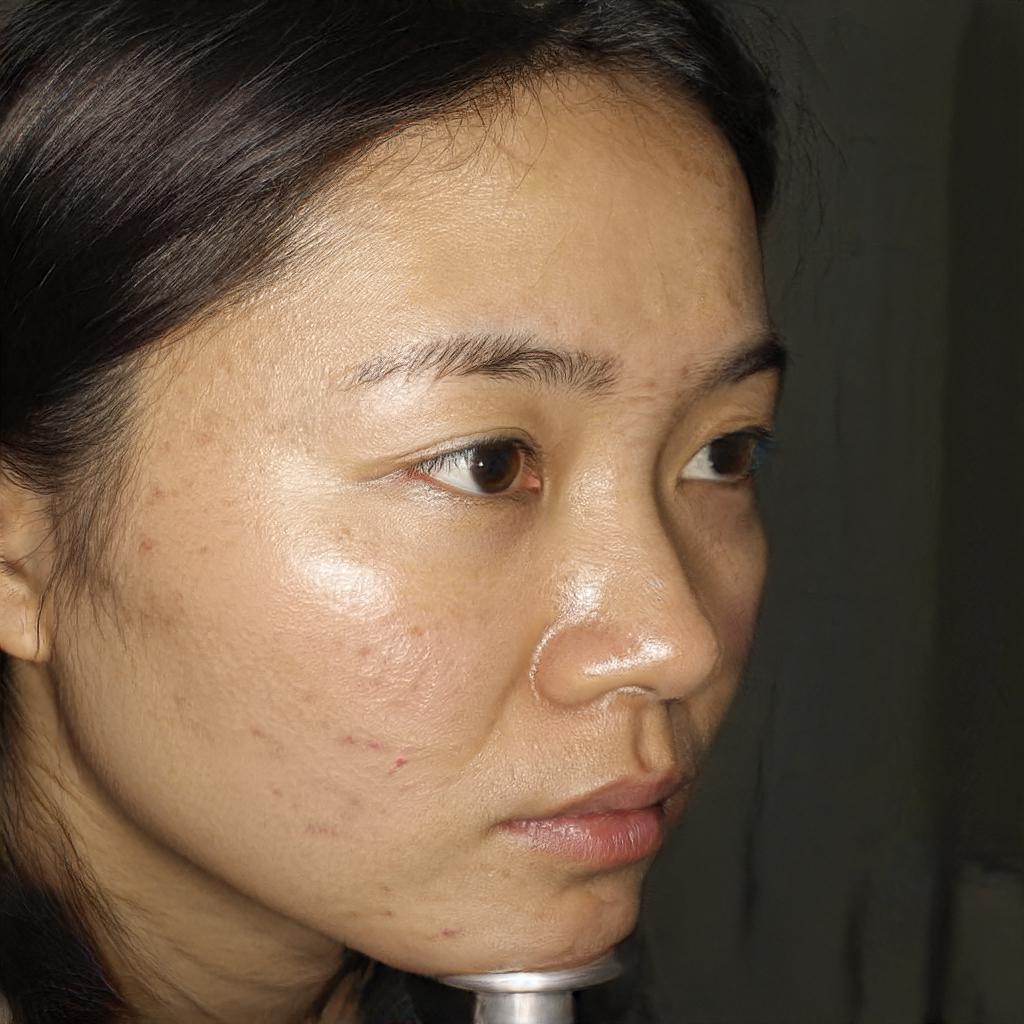}           & 
\includegraphics[height=3cm,width=3cm]{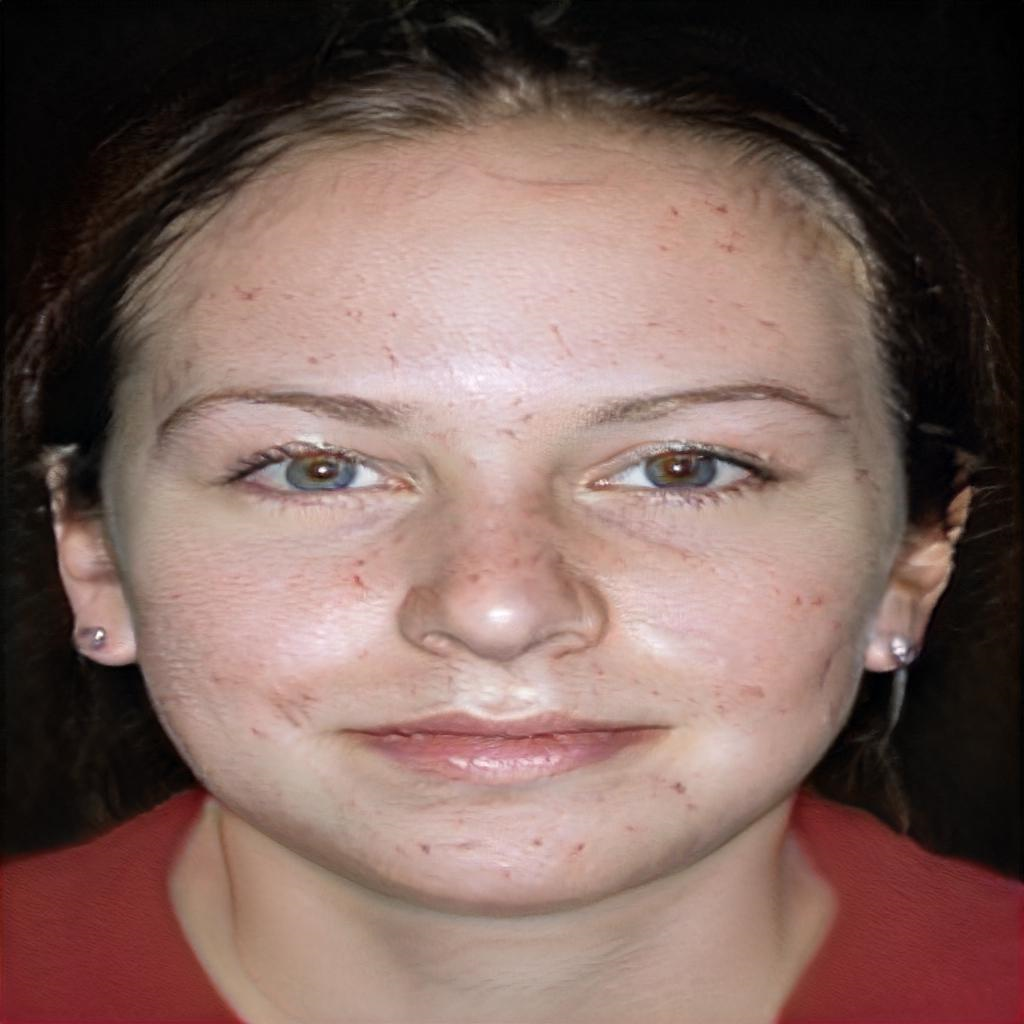}              & \includegraphics[height=3cm,width=3cm]{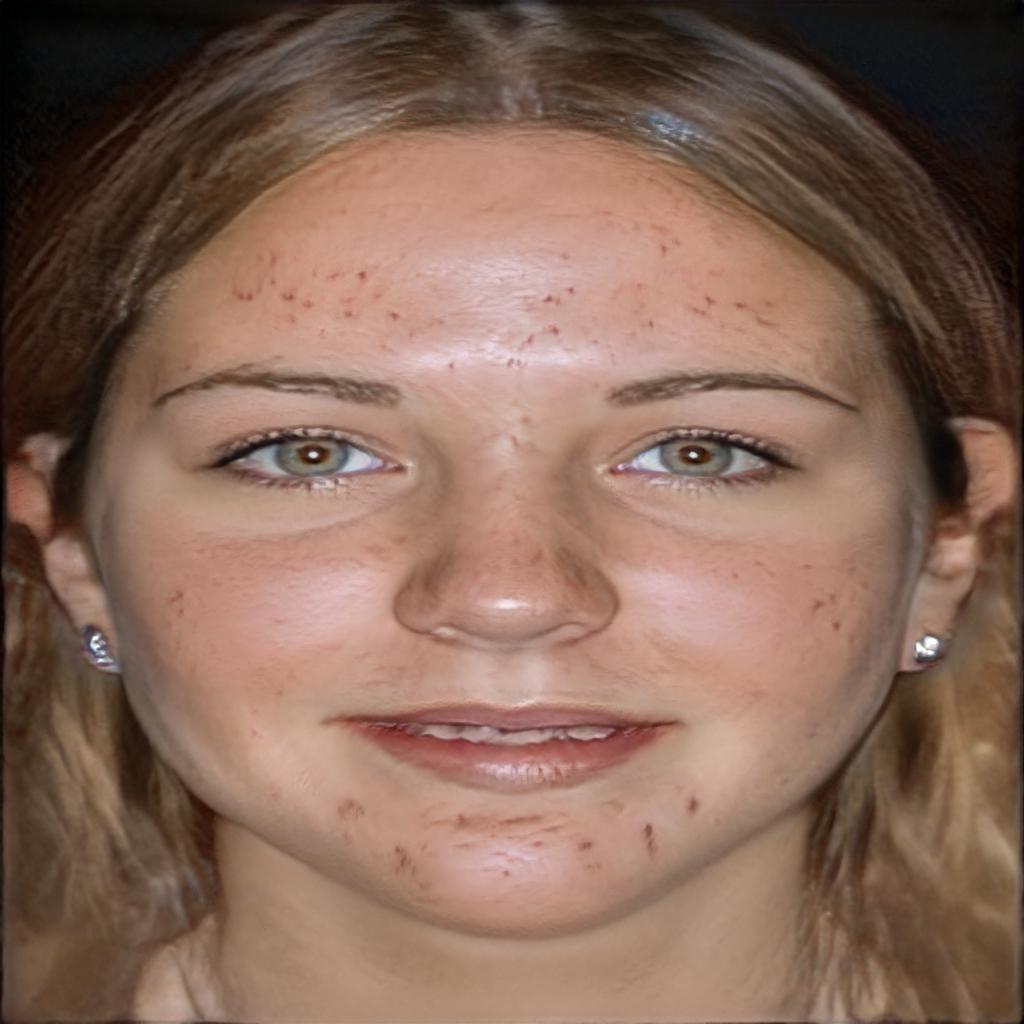}                     & \includegraphics[height=3cm,width=3cm]{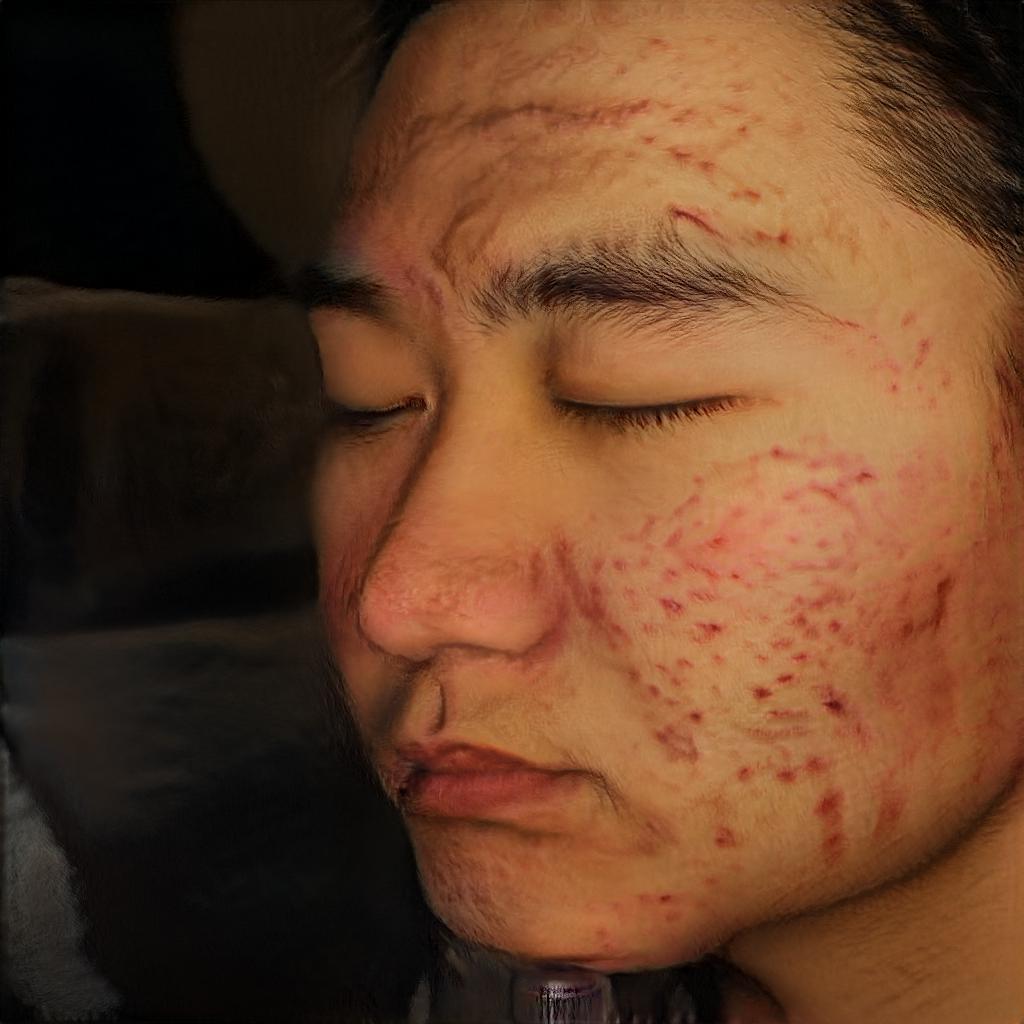}                  \\ 

\includegraphics[height=3cm,width=3cm]{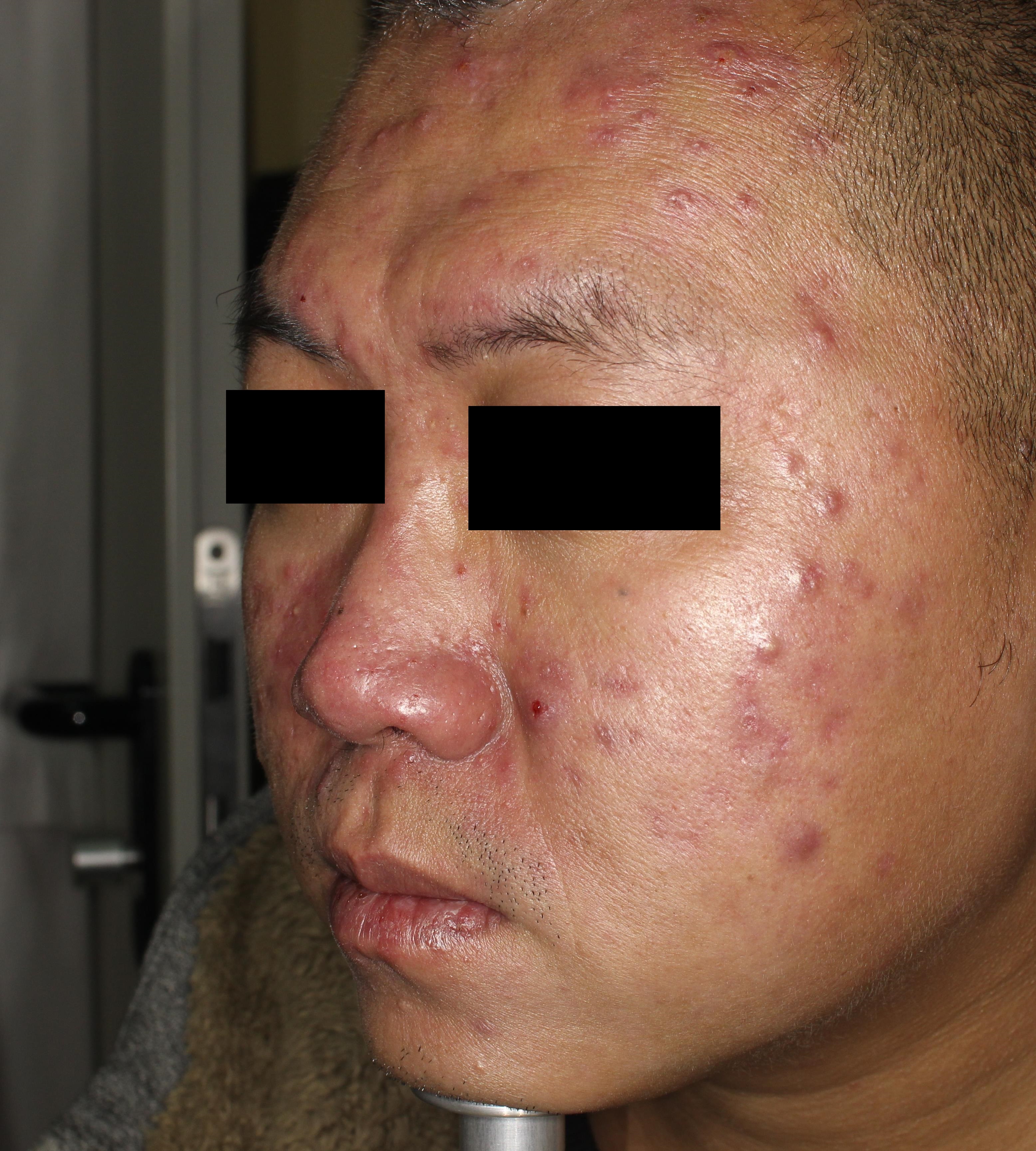}             & \includegraphics[height=3cm,width=3cm]{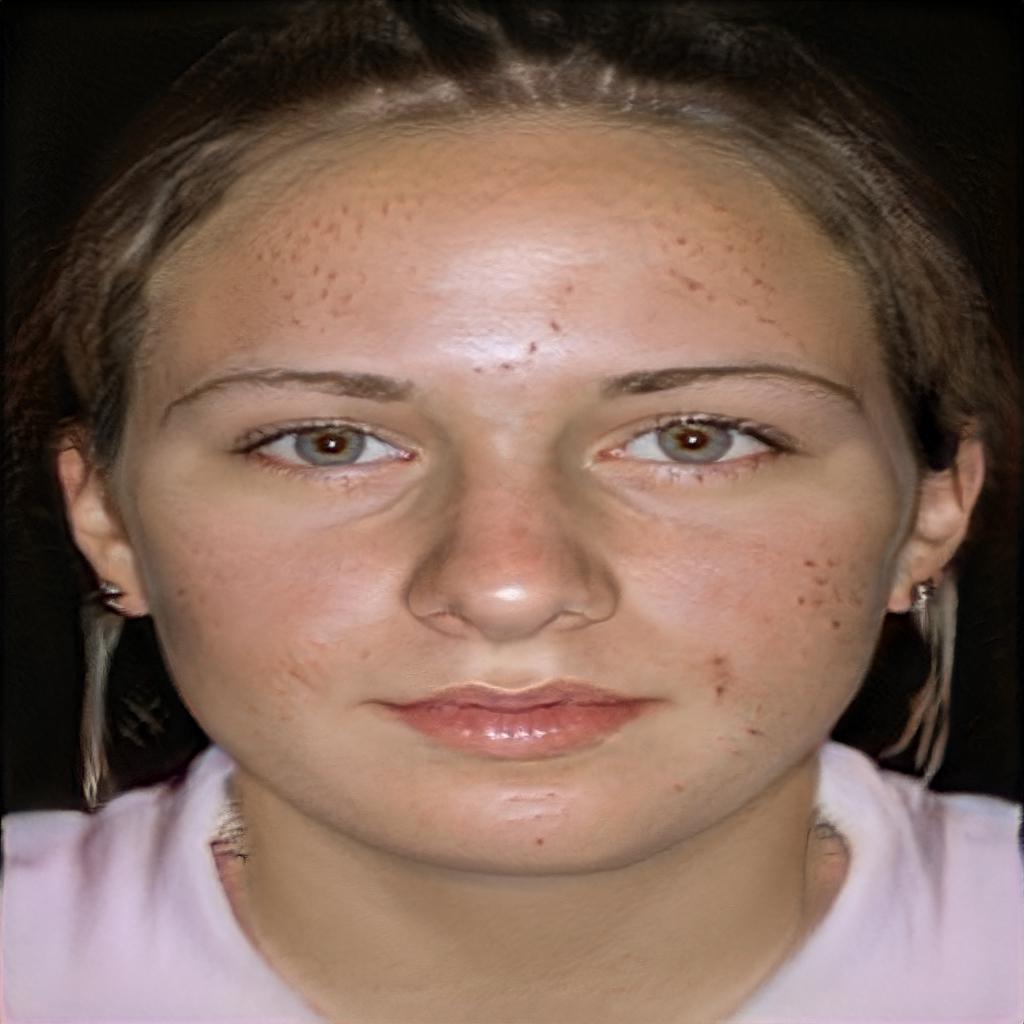}             & \includegraphics[height=3cm,width=3cm]{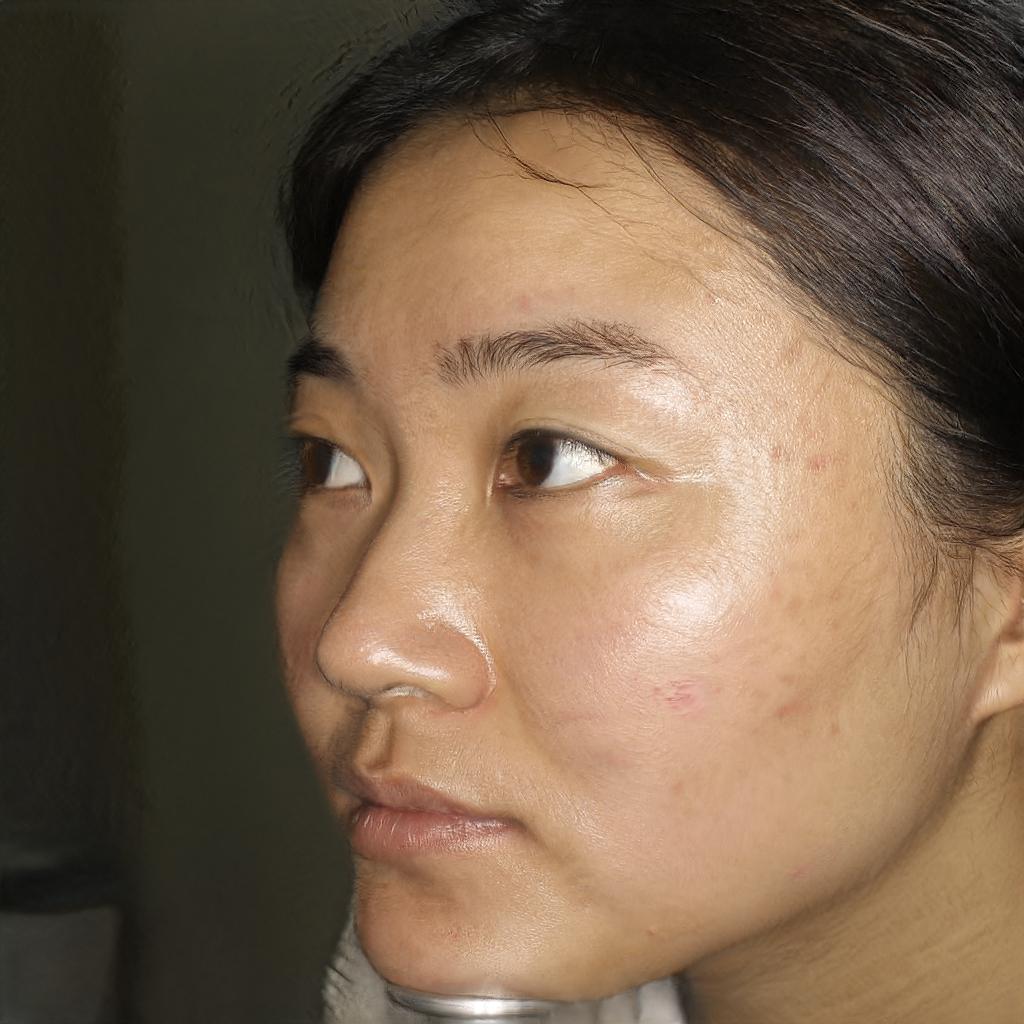}              & \includegraphics[height=3cm,width=3cm]{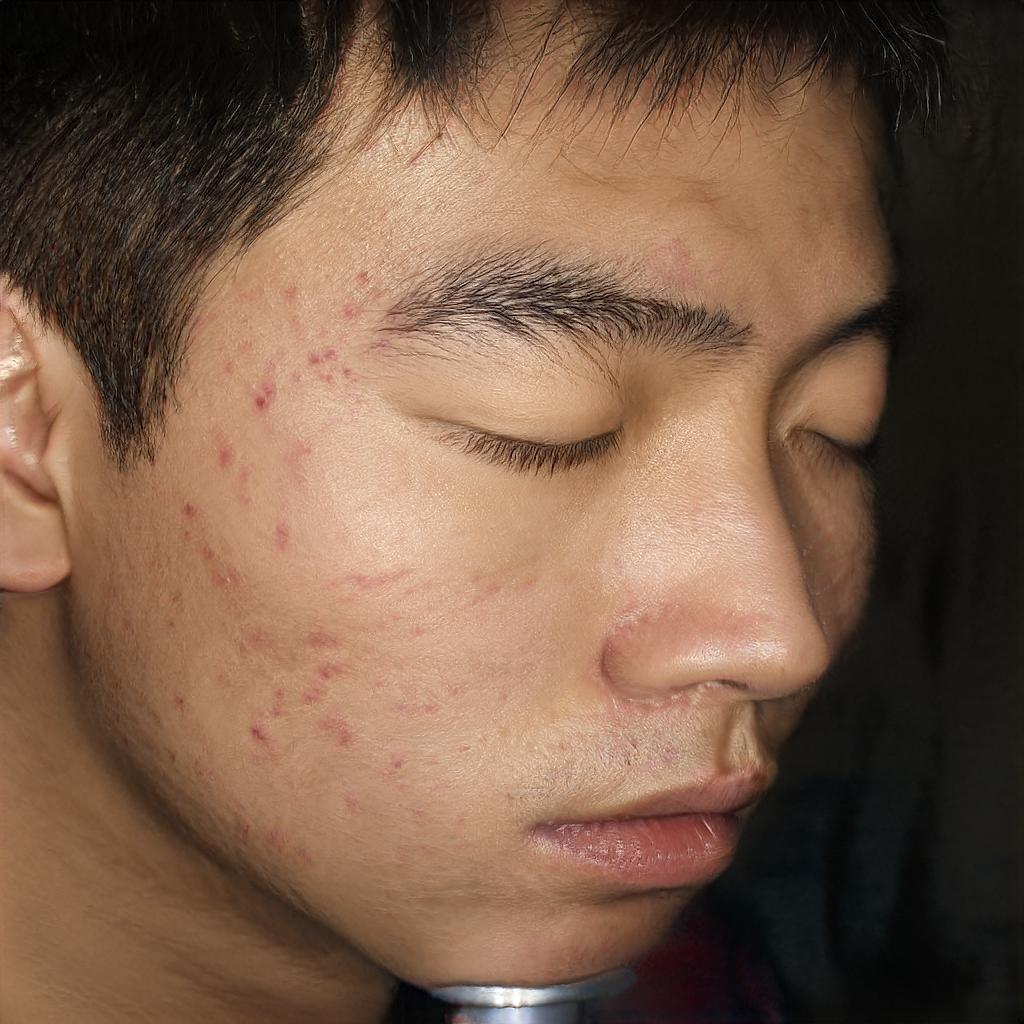}                    & \includegraphics[height=3cm,width=3cm]{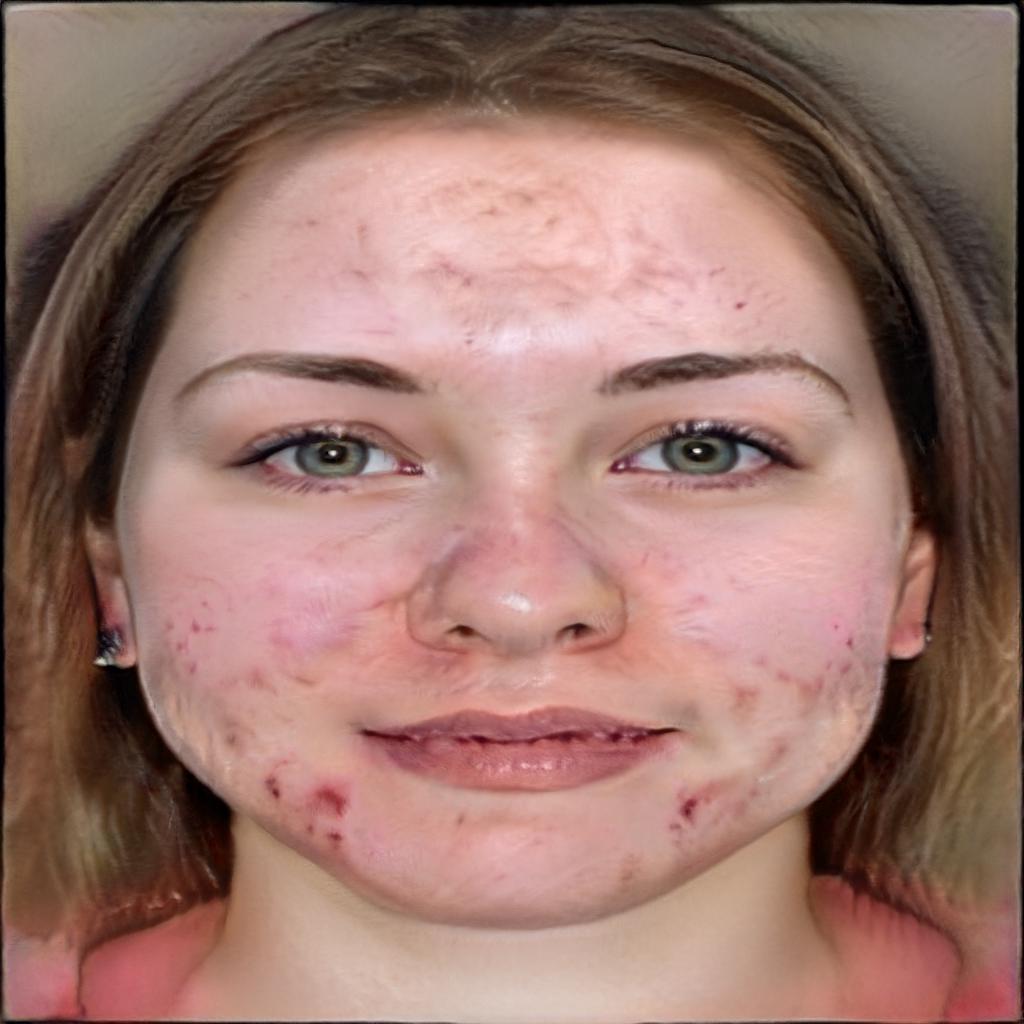}                   \\ 

\includegraphics[height=3cm,width=3cm]{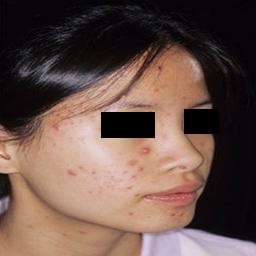}            & 
\includegraphics[height=3cm,width=3cm]{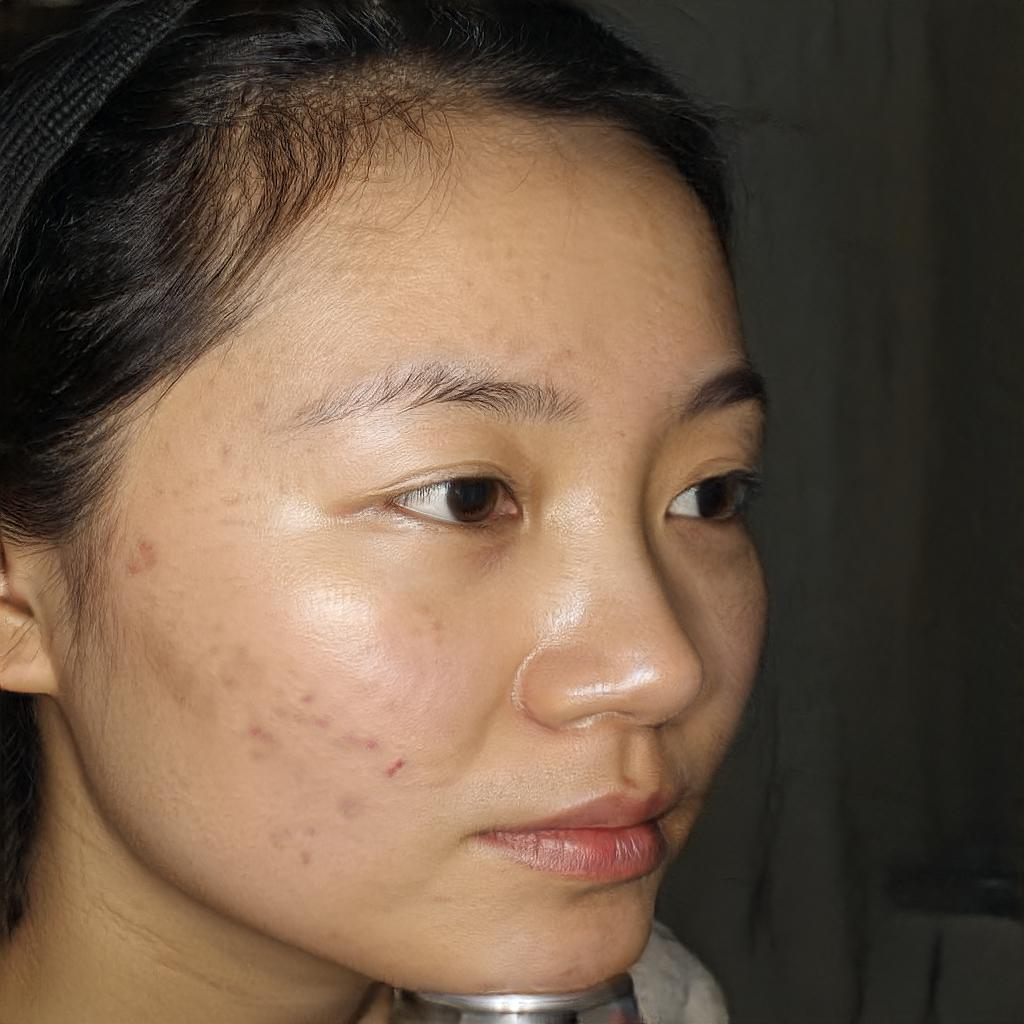}             & \includegraphics[height=3cm,width=3cm]{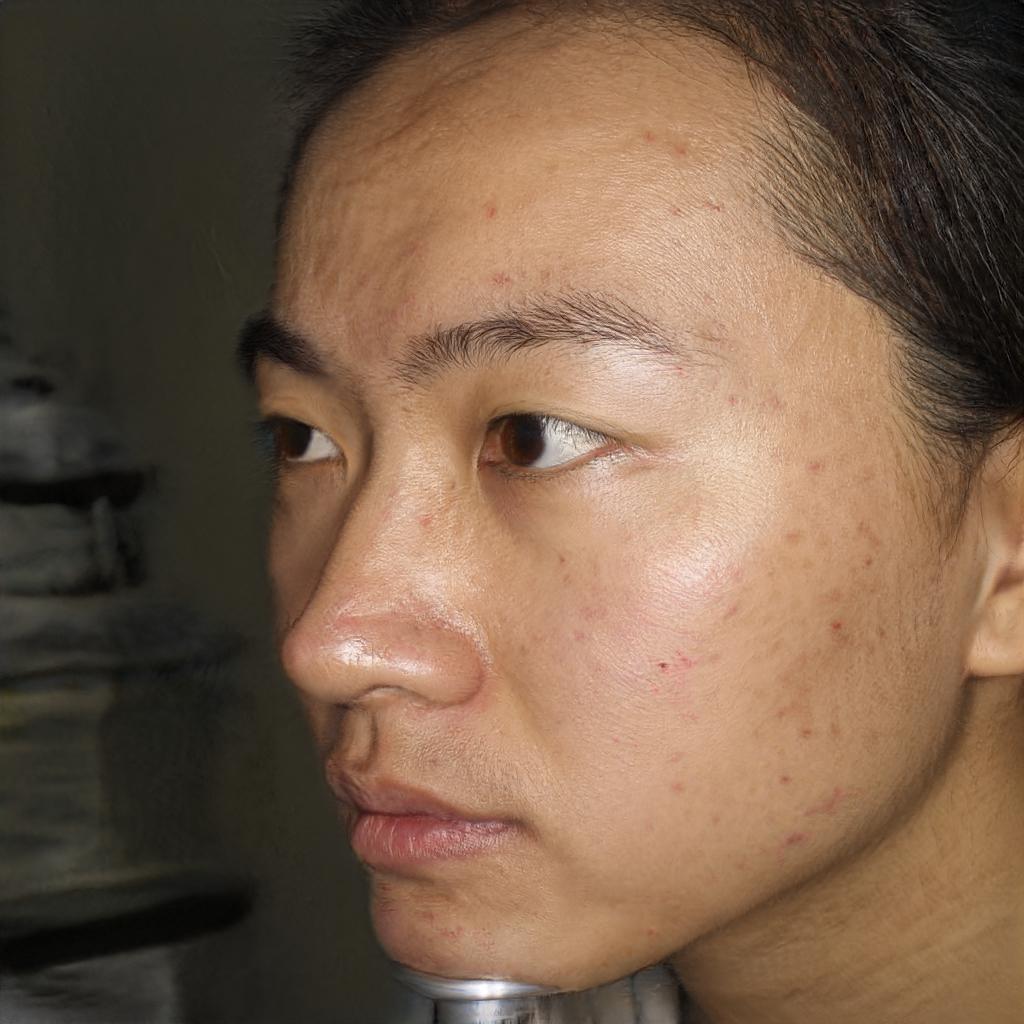}               & 
\includegraphics[height=3cm,width=3cm]{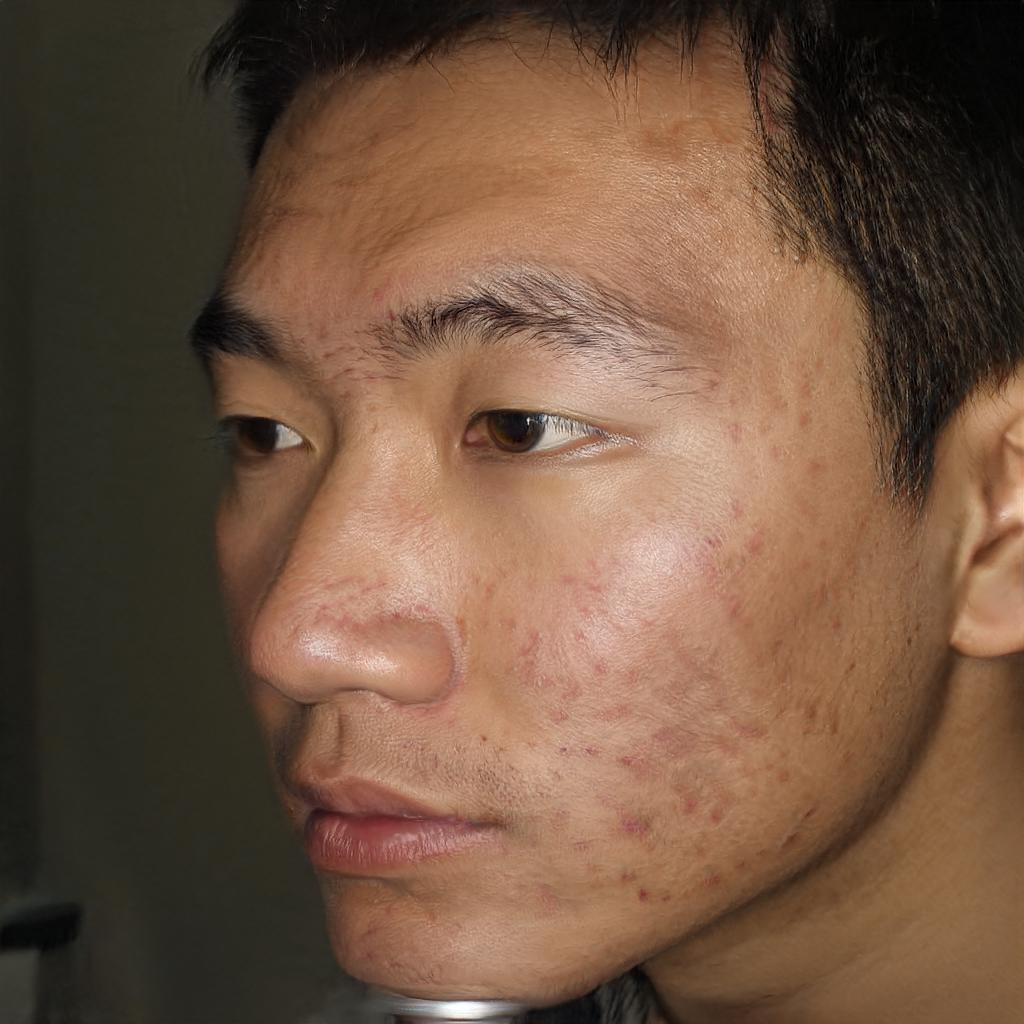}                           & 
\includegraphics[height=3cm,width=3cm]{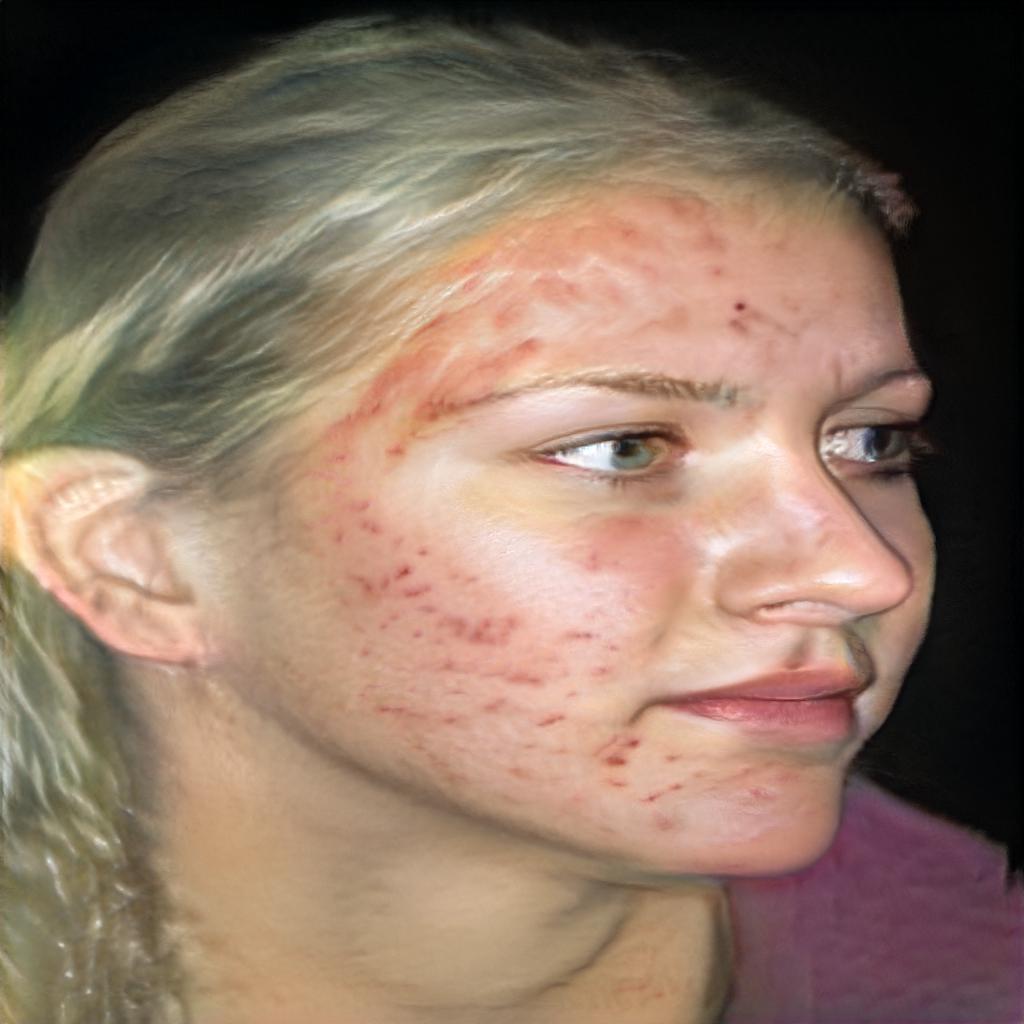}                   \\ 
\small{(a)}         & \small{(b)}         & \small{(c)}             & \small{(d)}                 & \small{(e)}               \\ 
\end{tabular}
}

\caption{Samples from the real dataset and each model's output : (a) Samples from the gathered dataset (b) Images generated from the first model (c) Images generated from mild acne model (d) Images generated from moderate acne model (e) Images generated from severe acne model}
\label{tab:res-images}
\end{figure}

After we finished training the first facial acne model and gathered 3 datasets of mild, moderate and severe facial acne images for training three additional StyleGAN2 models. We reached the results shown in figure \ref{tab:GAN-GRAPHS}, each graph represents the FID of the model at each tick. first facial acne model (a) trained on 1473 real facial acne images, started with an FID of 115.1972 at tick 0 reaching the lowest of 7.209 at tick 70, training took 5d 17h 53m. Mild Acne Model (b) trained on 1500 generate facial acne images, started with an FID of 91.2631 reaching the lowest of 16.0271 at tick 30, training took 1d 18h 39m. moderate acne model (c) trained on 1502 synthetic images, was the least performing model reaching the lowest FID of 108.6823, training took 4d 13h 52m. Severe Acne Model (d) trained on 1514 synthetic images, started with and FID of 102.1046 reaching the lowest FID of 11.1161 at tick 70, training took 4d 00h 47m. Figure \ref{tab:res-images} contains real images from the first dataset and samples generated by each StyleGAN2 model.


\begin{table}[H]
\caption{Parameters used in training StyleGAN2 model, during training a snapshot of the model is saved every 10 ticks.}
\centering
\def\arraystretch{1.5}
\begin{tabular}{|c|c|}
\hline
\textbf{Parameter}       & \textbf{Value} \\ \hline
image\_snapshot\_ticks   & 10             \\ \hline
mirror\_augment          & True           \\ \hline
network\_snapshot\_ticks & 10             \\ \hline
D\_lrate\_base           & 0.002          \\ \hline
G\_lrate\_base           & 0.002          \\ \hline
minibatch\_gpu\_base     & 4              \\ \hline
minibatch\_size\_base    & 32             \\ \hline
rnd.np\_random\_seed     & 1000           \\ \hline
total\_kimg              & 25000          \\ \hline
\end{tabular}
\label{tab:my-table3}
\end{table}

\subsection{CNN Models Performance Evaluation}
For evaluating our CNN models we used the following metrics: accuracy, precision, recall and F1 score. Accuracy is the ratio of correctly classified samples over the total number of samples. Precision is the fraction of correctly classified positive samples over the total of positive samples. Recall is the fraction of positive samples over the total of correctly classified samples. F1 score is the harmonic mean between precision and recall \cite{b17}.

\begin{equation} 
Accuracy = \frac{TP + TN}{TP + TN + FP + FN} 
\label{eq2}
\end{equation}

\begin{equation} 
Precision = \frac{TP}{TP + FP} 
\label{eq3}
\end{equation}

\begin{equation} 
Recall = \frac{TP}{TP + FN} 
\label{eq4}
\end{equation}

\begin{equation} 
F1 = 2\times\frac{Precision\times Recall}{Precision + Recall} 
\label{eq5}
\end{equation}

\begin{figure}[H]
\centering
\resizebox{0.91\textwidth}{!}{
\begin{tabular}{c  c  c c}
\includegraphics[width=0.5\textwidth]{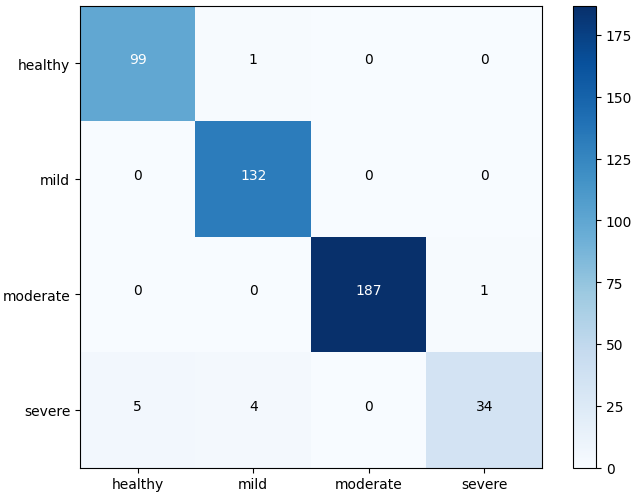}
 & \includegraphics[width=0.5\textwidth]{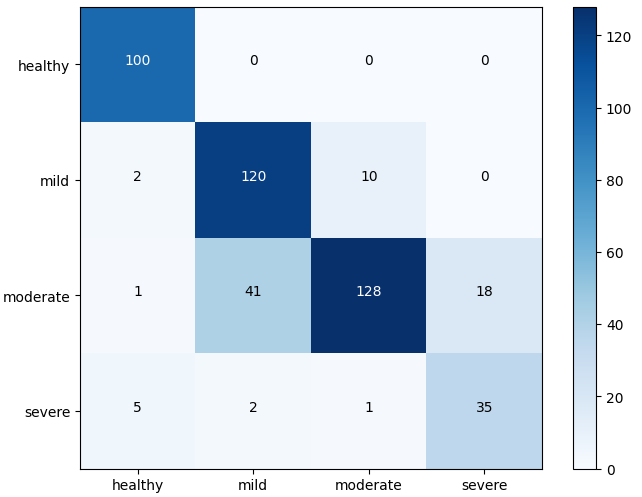} &
\includegraphics[width=0.5\textwidth]{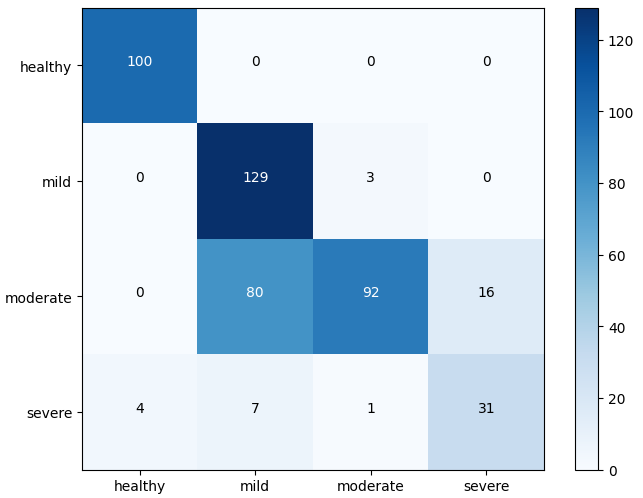} &                   \\ 
\large{(a)} & \large{(b)} & \large{(c)} \\
\end{tabular}
}

\caption{Confusion matrix for each CNN model tested on unseen images: (a) Confusion Matrix of InceptionResNetV2: ACCRUACY: 97.6\% , (b) Confusion Matrix of ResNet152V2: ACCRUACY: 82.7\% , (c) Confusion Matrix of ResNet50V2: ACCRUACY: 76.03\%.}
\label{tab:unseen}
\end{figure}

\begin{figure}[H]
\centering
\begin{minipage}{.45\linewidth}
  \includegraphics[width=\linewidth]{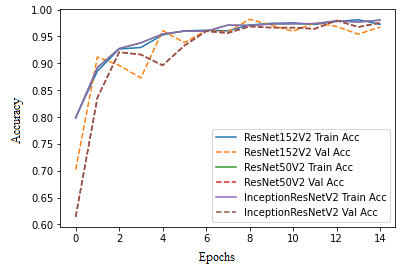}
  \caption{InceptionResNetV2, ResNet152V2 and ResNet50V2 models train and validation accuracy graphs}
  \label{acc-graph}
\end{minipage}
\hspace{.05\linewidth}
\begin{minipage}{.45\linewidth}
  \includegraphics[width=\linewidth]{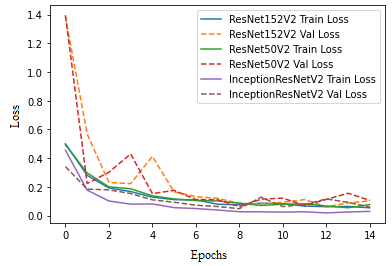}
  \caption{InceptionResNetV2, ResNet152V2 and ResNet50V2 models train and validation loss graphs}
  \label{loss-graph}
\end{minipage}
\end{figure}

In this paper we trained three CNN models using transfer learning on 4 classes of facial acne severity images using only synthetic images. The dataset was comprised of: 1387 healthy images, 841 mild Images, 1173 moderate images and 1086 severe images. InceptionResNetV2 Performed the best with an accuracy of 98.44\%, precision 98.33\% , 98.19\% recall and 98.25\% F1 score as shown in table \ref{tab:CNN-RES}. Both loss and accuracy curves in figures \ref{acc-graph} and \ref{loss-graph}  are converging together showing no signs of overfitting. These results are not enough to prove that these models can classify facial acne correctly in real scenarios, therefor the evaluation of each model was done using unseen real facial acne images. Figure \ref{tab:unseen} shows that only InceptionResNetV2 (a) had a high accuracy of 97.6\% and the other models struggled with classifying moderate class correctly. This proves that InceptionResNetV2 is able to classify facial acne images correctly in real scenarios and also synthetic images generated using GAN models can be used in deep learning applications and give same results as real images.

\begin{table}[H]
\caption{Parameters used in training CNN models. Model Checkpoint was used to save the model on each iteration if validation accuracy improves.}
\centering
\def\arraystretch{1.4}
\begin{tabular}{|c|c|}
\hline
\textbf{Parameter} & \textbf{Value}                      \\ \hline
Batch Size         & 16                                  \\ \hline
Epochs             & 15                                  \\ \hline
Optimizer          & ADAM                                \\ \hline
Learning Rate      & 0.0001                              \\ \hline
Classes            & 4                                   \\ \hline
Total Images       & 4487                                \\ \hline
Train/Test Split   & 80\%/20\%                           \\ \hline
IMG\_SIZE          & 224x224                             \\ \hline
Augmentation       & Rotation/Brightness/Horizontal Flip \\ \hline
Model Checkpoint   & True                                \\ \hline
\end{tabular}
\label{tab:my-table1}
\end{table}

\begin{table}[H]
\caption{CNN models performance evaluation using the following metrics: accuracy, precision, recall and F1 score.}
\centering
\def\arraystretch{1.4}

\begin{tabular}{|c|c|c|c|c|c}
\hline
\multicolumn{1}{|c|}{Model} & \multicolumn{1}{c|}{Accuracy} & Precision      & Recall & F1 Score         \\ \hline
\textbf{InceptionResNetV2}  & \textbf{98.44}                & \textbf{98.33} & \textbf{98.19} & \textbf{98.25} \\ \hline
ResNet152V2                 & 97.77                         & 97.61          & 97.38        & 97.48 \\ \hline
ResNet50V2                  & 97.88                         & 97.64          & 97.53        & 97.58  \\ \hline
\end{tabular}
\label{tab:CNN-RES}
\end{table}

\begin{figure}[H]
\centering
\resizebox{0.91\textwidth}{!}{
\begin{tabular}{cccc}
\multicolumn{1}{c}{\includegraphics[width=0.5\textwidth]{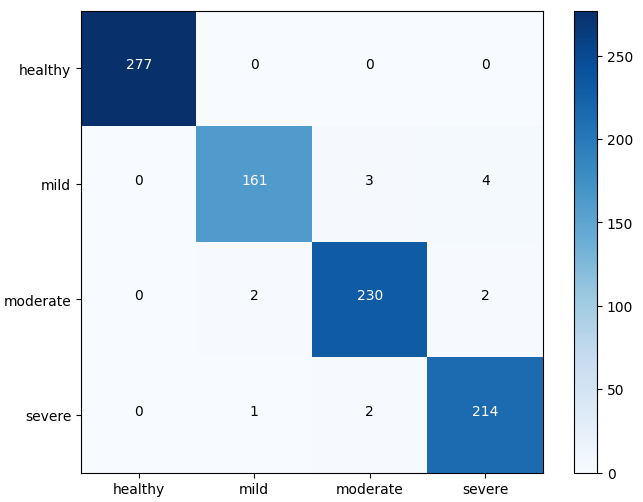}}
 & \multicolumn{1}{c}{\includegraphics[width=0.5\textwidth]{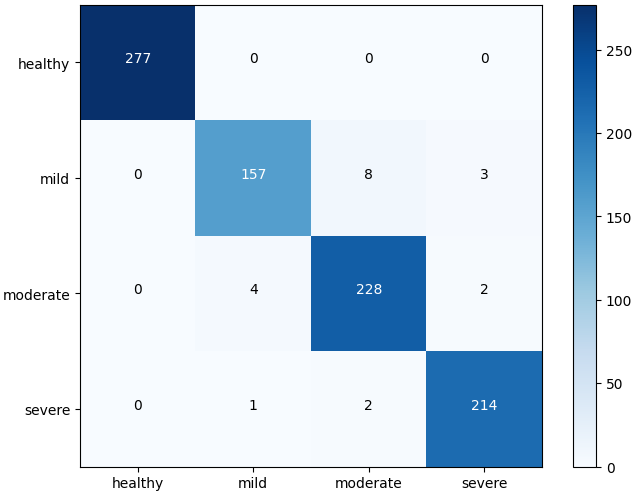}} &
\includegraphics[width=0.5\textwidth]{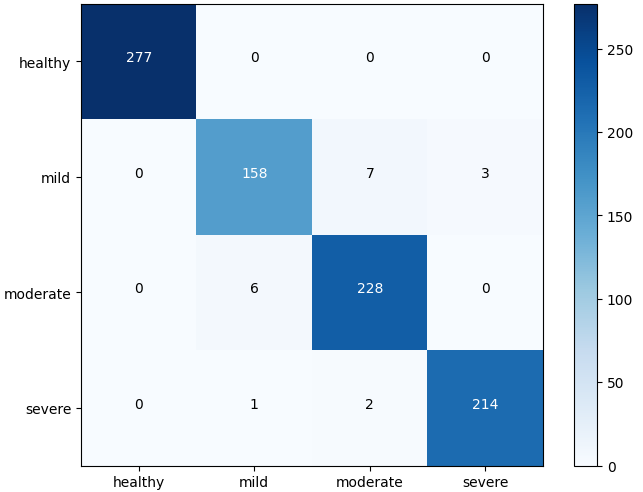} &                 \\
\large{(a)} & \large{(b)}& \large{(c)} \\
\end{tabular}
}
\caption{Confusion matrix for each CNN model on test split: (a) InceptionResNetV2 Confusion Matrix, (b) ResNet152V2 Confusion Matrix, (c) ResNet50V2 Confusion Matrix.}
\label{tab:my-table2}
\end{figure}

\section{Conclusion}
\label{conclusion}
This paper tackles the problem of the unavailability of free public dermatology datasets and their privacy. It offers a solution to help researchers and biomedical engineering communities to generate anonymous datasets for performance evaluation of their processing approaches. This study considered facial acne disease, an inflammatory skin disorder affecting around 9.4\% of the population \cite{conc}. In our experiment,  we collected 1473 images from multiple sources. These images were pre-processed (e.g. resized down or upscaled) using super-resolution deep-learning to generate 1024x1024 images and preserve the required information. A first StyleGAN2 model was trained using merged acneic face images corresponding to three levels of severity: mild, moderate and severe. Afterwards, at the next step,  specific StyleGAN2 Models were used to generate each separately large-size dataset. The last phase of this work is to prove that training classifiers using synthetic acneic face images does not affect the classification performance when authentic acneic face images were used in the test phase. Moreover,  we showed that the hybrid synthetic-authentic classifier based on  InceptionResNetV2 achieved the best accuracy of 97.6\%. 
Finally, the proposed approach described in this paper allows the biomedical engineering community to generate unlimited synthetic and realistic images of acneic faces without facing privacy issues or the lack of datasets. In addition, dermatologists can also use these synthetic face images for educational purposes. Further work is needed to include more skin diseases with multiple levels of severity.



\end{document}